# Presenting Punctuation

Michael White
CoGenTex, Inc.
840 Hanshaw Rd.
Ithaca, NY 14850 USA
*mike@cogentex.com*

## 1. Introduction

Until recently, punctuation has received very little attention in the linguistics and computational linguistics literature, even though punctuation marks are evidently among the most important structural elements in written language. Since the publication of Nunberg's (1990) monograph *The Linguistics of Punctuation*, however, punctuation has seen its stock begin to rise: spurred in part by Nunberg's ground-breaking work, a number of valuable inquiries have been subsequently undertaken, including Hovy and Arens (1991), Dale (1991), Pascual (1993), Jones (1994), and Briscoe (1994). Of these recent works, those on the synthesis side have focused on ways of formalizing the discourse functions of punctuation, whereas those on the analysis side have emphasized how punctuation can improve robust parsing.

Continuing this line of research, I investigate in this paper how Nunberg's approach to presenting punctuation (and other formatting devices) might be incorporated into NLG systems. Insofar as the present paper focuses on the proper syntactic treatment of punctuation, it differs from the above works in that it is the first to examine this issue from the generation perspective.

The paper is organized as follows. In section 2, I introduce the phenomena of interest and describe Nunberg's account of them. In section 3, I evaluate Nunberg's approach from an NLG perspective, discussing both its empirical adequacy and theoretical perspicuity. In section 4, I identify the ingredients of the present account, which elaborates and improves upon aspects of Nunberg's; then, I sketch how this account can be instantiated within the Meaning-Text framework of the CoGenTex core realizer. Finally, in section 5, I review the contributions of the paper as well as the further questions it raises.

## 2. The Phenomena and Nunberg's Account of Them

Nunberg discusses a wide variety of phenomena concerning the presentation of punctuation in his monograph; to these, I will add a few previously unexamined but closely related cases. Though wide-ranging, Nunberg's account of these phenomena is not given in great detail, and often relies on questionable assumptions (as should become clear below). Nevertheless, in order to simplify the exposition, I will take Nunberg's account at face value in this section, leaving aside questions of theoretical perspicuity and empirical adequacy.

## *2.1 Absorptions and Transpositions*

### 2.1.1 Point absorption

Nunberg calls commas, dashes, semi-colons, colons and periods *points*. According to Nunberg, two points are never allowed to appear in sequence. When two points happen to coincide, one is *absorbed* by the other, according to the 'strength' of the two points. The strength ordering Nunberg assumes is as follows: the comma is the weakest point, the period the strongest, and the dash, the semi-colon and the colon are inbetween, in that order.

Central to Nunberg's generalization is the assumption that with point absorptions, scope and word order do not matter, only adjacency. To illustrate this claim, Nunberg provides the following examples, where the underscore serves to indicate the site of the 'missing' point (Nunberg 1990, pp. 58-59; original example numbers retained):

(5.6) John left, apparently; Mary stayed.
(5.7) John told them the news, apparently: Mary had left.
(5.13) John left; apparently, Mary stayed.
(5.14) John gave us the news: apparently, no one was coming.

Now, in order to see that a comma is 'missing' in each of these cases, one must first assume (with Nunberg) that comma-delimited parentheticals are always 'base-generated' with commas on both ends. Although this idea is intuitively clear, Nunberg is not very specific as to how it is intended to be fleshed out, and in particular does not address the question as to whether there is ever an intermediate structure where the 'missing' comma does in fact appear. For illustrative purposes, however, it is convenient to assume that the pertinent comma is inserted and then deleted by the point absorption rule. Figure 1 illustrates how this might work for example (5.6):

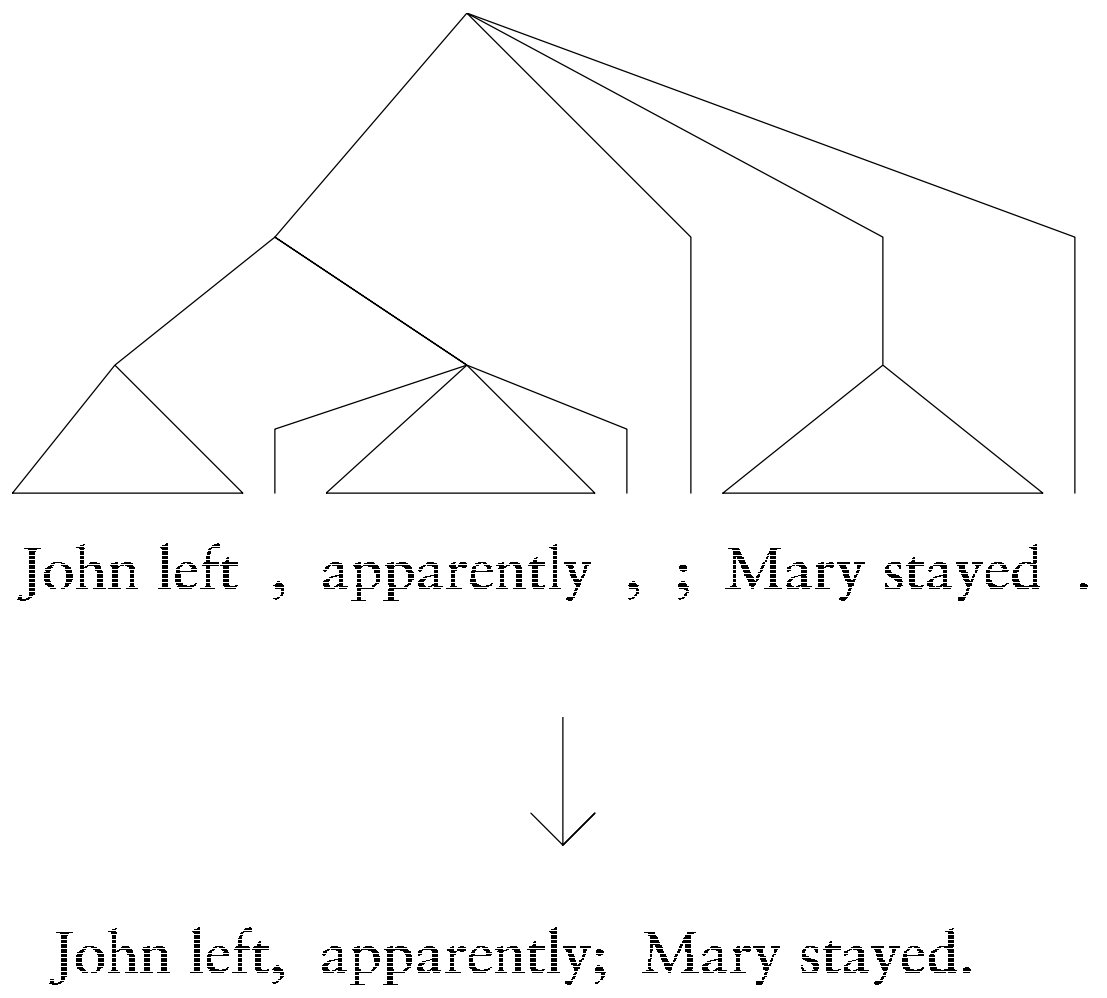

**Figure 1**

Nunberg's claim that the proper generalization concerns the absorption of adjacent points is further strengthened by examples where the 'missing' comma can be presumed to outscope the point that is actually presented, as in (5.11) below:

(5.11) Jones, who had never seen the picture—none of the crewmen had—was completely devastated.

Figure 2 illustrates the presumed syntax for the subject noun phrase:

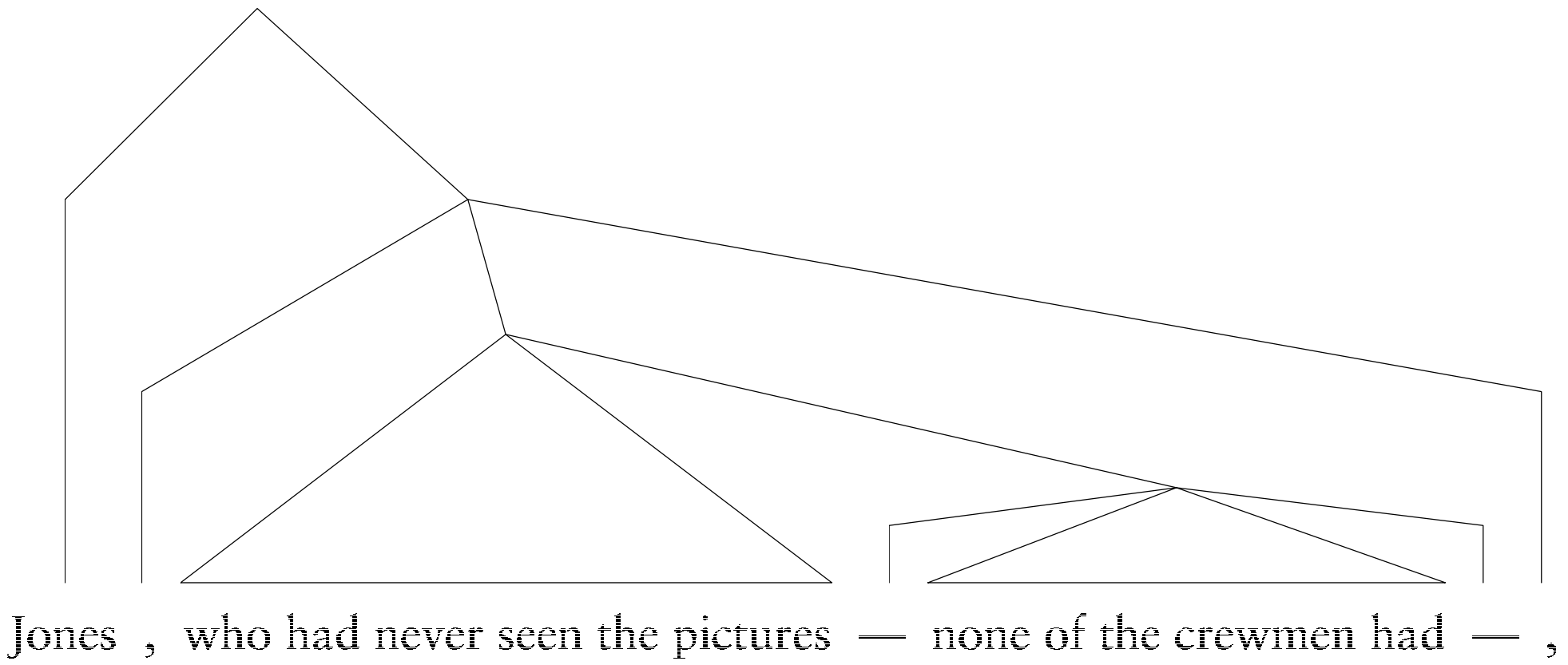

**Figure 2**

Interestingly, Nunberg claims that point absorption and many other punctuation phenomena involve tacit knowledge that is neither explicitly taught nor subject to typographer's convention. This view represents a rather bold departure from earlier ones, where most (if not all) punctuation phenomena were regarded as no more than arbitrary conventions subject to the whims of competing publishers. In support of his view, Nunberg provides numerous negative examples (some of which will be reviewed here) that reveal surprisingly strong intuitions about cases never discussed in style manuals. For instance, in a case similar to (5.11) above, examples (5.22) and (5.23) illustrate what happens when the 'missing' comma is presented rather than the dash it outscopes (p. 61):

(5.22) John will have to be asked to chair the first session—that would be ideal—or we will have to reschedule the entire conference.
(5.23) * John will have to be asked to chair the first session—that would be ideal, or we will have to reschedule the entire conference.

### 2.1.2 Bracket absorption

Nunberg uses the term *brackets* as a name for the class which includes parentheses and quotation marks in addition to brackets per se. What motivates Nunberg to give a name to this class is all of these marks behave alike with respect to absorption rules. While brackets absorb points, multiple brackets are never themselves absorbed; not surprisingly, brackets also exhibit scope differences, insofar as they only absorb points within their scope (p. 62):

(5.28) May failed the test (which was not surprising—she didn't study) and will have to repeat the course.
(5.29) May failed the test—which covered all the readings (including the book she had lost)—and will have to repeat the course.

A viable option not considered by Nunberg is that (5.28) involves a dash-expansion, analogous to a colon-expansion, rather than a dash-interpolated adjunct—i.e., that only a single dash is 'base-generated' in this example. For expository purposes, however, it is convenient to assume Nunberg's analysis is at least a possible one (we will return to this matter in section 3).

### 2.1.3 Quote Transposition

In American practice, end quotes are systematically transposed with commas and periods; however, exceptions do of course exist, especially in cases where it is important to be precise about that which is quoted (e.g. in computer manuals).

Interestingly, Nunberg demonstrates that both point absorption and bracket absorption must precede quote transposition. To see this, let us first consider why point absorption must precede quote transposition. Assuming the underlying (or 'base-generated') structure in (5.68), ordering these rules in the opposite way would yield (5.70), in which the comma fails to get absorbed by the semi-colon (p. 71):

(5.68) She came in her truck, which she called "li'l red",; it was a blue Toyota.
(5.70) * She came in her truck, which she called "li'l red,"; it was a blue Toyota.

Now, one might suspect that (5.70) could be saved by bracket absorption; but, there is independent evidence demonstrating that bracket absorption must also precede quote transposition. If we assume the underlying structure in (5.73), then ordering quote transposition before bracket absorption would yield (5.74), where the comma has been mistakenly deleted (p. 72):

(5.73) "Sentimentality", Wilde said, "is the attribution of tenderness to nature where God did not put it."
(5.74) * "Sentimentality" Wilde said, "is the attribution of tenderness to nature where God did not put it."

### 2.1.4 Graphic Absorption

This case concerns the absorption of sentence-ending periods—but not other points—by both abbreviation periods and tone indicators (such as question marks and exclamation points). Presumably, this phenomenon is a result of the graphical similarity between these marks, whence the name. The following examples illustrate the phenomenon (p. 67):

(5.58) He lives in D.C.
(5.61) He lives in D.C.; she lives in N.Y.
(5.54) I asked her, "Who is she?"
(5.62) Are you going to D.C.?

In regard to example (5.54), Nunberg suggests that quote transposition must apply before graphic absorption, as otherwise the period and the question mark would not be adjacent. In support of this assumption, Nunberg observes that this ordering accounts for the following examples without further ado (p. 67):

(5.52) I am puzzled (who is he?).
(5.53) I am annoyed (he is a fool!).
(5.54) I asked her, "Who is she?"
(5.55) I told her, "He is a fool!"

Examples (5.52) and (5.53) illustrate how intervening parentheses block graphic absorption. A matter not considered by Nunberg is whether font- and face-changes can likewise block graphic absorption; if one considers the formatting directives used in most markup languages to be on the same level as punctuation marks, then one would expect this to be the case. As it turns out though, such directives do not block graphic absorption, which strongly suggests that font- and face-changes are 'transparent' to adjacency.

To illustrate, let us consider example 5.5 from *The Chicago Manual of Style* (henceforth CMS, p. 159; original example numbers and formatting retained):

**5.5** After she wrote *What Next?*

Note that no period is presented here, despite the fact that this phrase constitutes a complete utterance (note also that this example was originally used to illustrate a separate phenomenon, as will be seen in section 2.2.3 below). In order for this case to be covered by graphic absorption, however, the question mark and period would have to be adjacent, which they would not be with most markup languages:[1]

(1) `*After she wrote :BEG-ITAL What Next? :END-ITAL .`

### *2.2 Alternations, Expansions, Harmonies and Promotions*

#### 2.2.1 Bracket and italics alternation

According to The University of Chicago Press guidelines, brackets should alternate to show embeddings (CMS, pp. 190, 365, resp.):

**5.129** During a prolonged visit to Australia, Glueck and an assistant (James Green, who was later to make his own study of a flightless bird [the kiwi] in New Zealand) spent several difficult months observing the survival behavior of cassowaries and emus.

**10.26** "Don't be absurd!" said Henry. "To say that 'I mean what I say' is the same as 'I say what I mean' is to be as confused as Alice at the Mad Hatter's tea party. You remember what the Hatter said to her: 'Not the same thing a bit! Why you might just as well say that "I see what I eat" is the same thing as "I eat what I see"!' "

Similar to bracket alternation is the one between italics and roman style, well-known from the LaTeX \em command.

#### 2.2.2 Colon- and dash-expansions

Curiously, colon-expansions are unlike bracket and italics alternation in that they resist multiple levels of embedding, as Nunberg points out (p. 31):

(4.23) * They serve a lot of cajun dishes: blackened redfish, gumbo and one thing you don't see a lot o<u>f: c</u>atfish sushi.

Note that this problem can be avoided by using a dash-expansion instead (a possibility not considered by Nunberg):

(2) They serve a lot of cajun dishes: blackened redfish, gumbo and one thing you don't see a lot o<u>f—c</u>atfish sushi.

In a similar vein, Nunberg observes that dash-interpolated adjuncts can be used to avoid nesting parentheses (p. 35):

[1] Nunberg suggests that standard formatting directives might turn out to be unappealing from a linguistic standpoint, as seems to be the case here, though he does not pursue the matter.

(4.45) * Jones (an employee (actually, a director) of the firm) was also present.
(4.46) Jones (an employee—actually, a director—of the firm) was also present.

### 2.2.3 Font- and face-harmony

In general, The University of Chicago Press suggests setting punctuation marks in the same font and face as the adjacent text. For example, in 5.4 below, the semi-colon and colon are recommended to appear in italics and bold, respectively (p. 158):

**5.4** Luke 4:16*a;*
**Point:** one-twelfth of a pica

As with quote transposition, though, there are exceptions to this rule. For example, with tone indicators immediately following italicized titles, The University of Chicago Press recommends against setting these in italics in order to avoid misreading (p. 159):

**5.5** When did she write *Together Again*?
After she wrote *What Next?*

In the first, interrogative sentence, the question mark is not set in italics, in order to avoid misreading the title; in contrast, in the second, declarative sentence, the question mark is set in italics, since it does form part of the title. In other words, in these cases italics does not spread beyond its logical scope.

A case considered neither by Nunberg nor by The University of Chicago Press is the following one, suggested by Igor Mel'cuk:

(3) When did she write *What Next?*

This example suggests that graphic absorption might also apply to tone indicators, though perhaps an additional question mark (not set in italics) would also be acceptable here.

Another interesting case of font- and face-harmony concerns brackets. With these, The University of Chicago Press recommends harmony only when it is consistent at both ends (p. 159):

**5.6** *(express violations)*
[it was substituted for *outrageous*]

As with the other exceptional cases, this recommendation is an instance of structural considerations impinging on graphically motivated formatting decisions.

### 2.2.4 Comma promotion

The University of Chicago Press recommends promoting commas to semi-colons when items in a series involve 'internal punctuation'—which presumably includes at least commas and dashes (CMS, p. 182). Note that this holds irrespective of whether these items are clauses or simpler phrases:[2]

**5.94** The membership of the international commission was as follows: France, 4; Germany, 5; Great Britain, 1; Italy, 3; the United States, 7.

As Nunberg points out, comma promotion must precede the absorptions and transpositions described above. For example, in the case of quote transposition, whether a list separator is transposed depends on how it is realized (p. 65):

(5.43) He repeated all the great naval mottoes: Nelson's "England expects that every man will do his duty," John Paul Jones' "I have not yet begun to fight," and …
(5.44) He repeated all the great naval mottoes: Nelson's "England expects that every man will do his duty," which is actually a misquotation; John Paul Jones' "I have not yet begun to fight"; ...

As another example, consider the case of point absorption. Here Nunberg argues that since dashes come between commas and semi-colons in the absorption ordering, one cannot determine whether to leave or delete a delimiter dash until the type of adjacent separator (comma or semi-colon) has been determined, citing the following examples as evidence (p. 61):

(5.22) John will have to be asked to chair the first session—that would be ideal—or we will have to reschedule the entire conference.
(5.24) Either the opening remarks will have to be kept to fifteen minutes, or John will have to be asked to chair the first session—that would be ideal; or we will have to reschedule the entire conference.

(While I agree with Nunberg that retaining the semi-colon over the dash is preferable in the second example, I still find it somewhat marginal.)

## *2.3 Discourse-Level Punctuation*

### 2.3.1 Brackets revisited

Though Nunberg does not emphasize this point, it is important to note that bracket absorption only applies sentence-internally. To see this, cf. the following pairs of examples (CMS, pp. 159, 162):

**5.8** The snow (she caught a glimpse of it as she passed the window) was now falling heavily.

[2] Nunberg (p. 45) notes two further complexities to this phenomenon that will not be addressed here—namely, that comma promotion is suppressed when only the last item in a series contains 'internal punctuation,' and also that it does not occur beyond the highest level separators.

Gilford's reply, "He appears to be untrustworthy," was unexpected.

**5.14** Florelli insisted on rewriting the paragraph. (I had encountered this intransigence on another occasion.)

"She was determined never again to speak to him [Axelrod]."

That the difference between 5.8 and 5.14 concerns the relative scope of punctuation and sentence boundaries is further clarified by the following passage (used here as an example) and example from CMS (pp. 190, 367, resp.):

**5.127** Commas, semicolons, colons, and dashes should be dropped before a closing parenthesis. Such punctuation, moreover, should not be used before an opening parenthesis unless the parentheses are used to mark divisions or enumerations run into the text (see 5.126). If required by the context, other nonterminal punctuation should follow the closing parenthesis. (For more regarding the use of other punctuation with parentheses, and for examples, see under individual marks: 5.14, 5.20, 5.28. For use of the single parenthesis with figures and letters in outline style see 8.79.)

**10.36** "Ransomed? What's that?"
"I don't know. But that's what they do. I've seen it in books; and so of course that's what we've got to do."

### 2.3.2 Vertical lists

Enumerated and bulleted lists often contain multiple sentences. When the list items do form a single sentence, however, such lists can be punctuated as if they were run-in rather than vertical (p. 160):[3]

**5.10** After careful investigation the committee was convinced that
1. the organization's lawyer, Watson, had consulted no one before making the decision;
2. the chair, Fitcheu-Braun, had never spoken to Watson;
3. Fitcheu-Braun was as surprised as anyone by what happened.

In particular, note here that with the enumeration folded into the sentence comma promotion applies, resulting in the use of semi-colons as list separators.

## 3. Nunberg's Approach from an NLG Perspective

Though generally admired, the account set forth in Nunberg's monograph has been criticized on both empirical and theoretical grounds. On the empirical side, Sampson (1993) has challenged some of Nunberg's claims as empirically inadequate; specifically, he has shown that (1) Nunberg is mistaken to assert that there is a simple, dialectal difference between American and British quotation practice, and (2) there are counter-examples to Nunberg's claims concerning syntactic restrictions on colon-expansions. Because these aspects of Nunberg's account fall outside the main focus of the present study, I will not dwell on this issue further here.

[3] Note that the conjunction preceding the final item is optional with vertical lists.

More important for present purposes are Briscoe's (1994) criticisms of the theoretical perspicuity of Nunberg's account. Briscoe questions Nunberg's assumption that all dash-interpolated adjuncts are underlyingly balanced, which necessitates the invocation of the point absorption rule to convert (4) to (5) below (p. 7, example numbers changed):

(4) * Max fell—John had kicked him—.
(5) Max fell—John had kicked him.

In cases such as this one, a dash-expansion—where a single dash separates the two clauses, rather than having two dashes delimit the second clause—would seem to be as much if not more natural than Nunberg's analysis.

Briscoe uses this case in making the following arguments against Nunberg's approach to presenting punctuation:

> The various rules of absorption introduce procedurality into the grammatical framework and require the positing of underlying forms which are not attested in text. For this reason, I make no use of such rules but rather capture their effects through propagation of featural constraints in parse trees. For instance, (4) is blocked by including distinct rules for the introduction of balanced and unbalanced text adjuncts and only licensing the latter sentence finally.

While Nunberg's account does appear rather baroque when one focuses on this particular case, it should be noted that Briscoe does not address those cases where Nunberg's account shines. In particular, if one recalls Figure 2, which accompanies example (5.11), it should be evident that in cases such as these relying on adjacency yields a much simpler story than relying on hierarchy. This suggests that Nunberg's general approach, if not applied overzealously, has the potential to considerably simplify the problem of properly presenting punctuation.

To cite another case, consider (6) and (7) below:

(6) During the month of April, the preliminary design of Project Reporter continued.
(7) During the month of April—the last month with available data—the preliminary design of Project Reporter continued.

Example (6) is taken from CoGenTex's Project Reporter (Korelsky et al., 1993) test corpus; example (7) is an invented variant. Borrowing the best from both Briscoe's and Nunberg's approaches, example (7) can be treated as shown by the illustrative syntax in Figure 3:

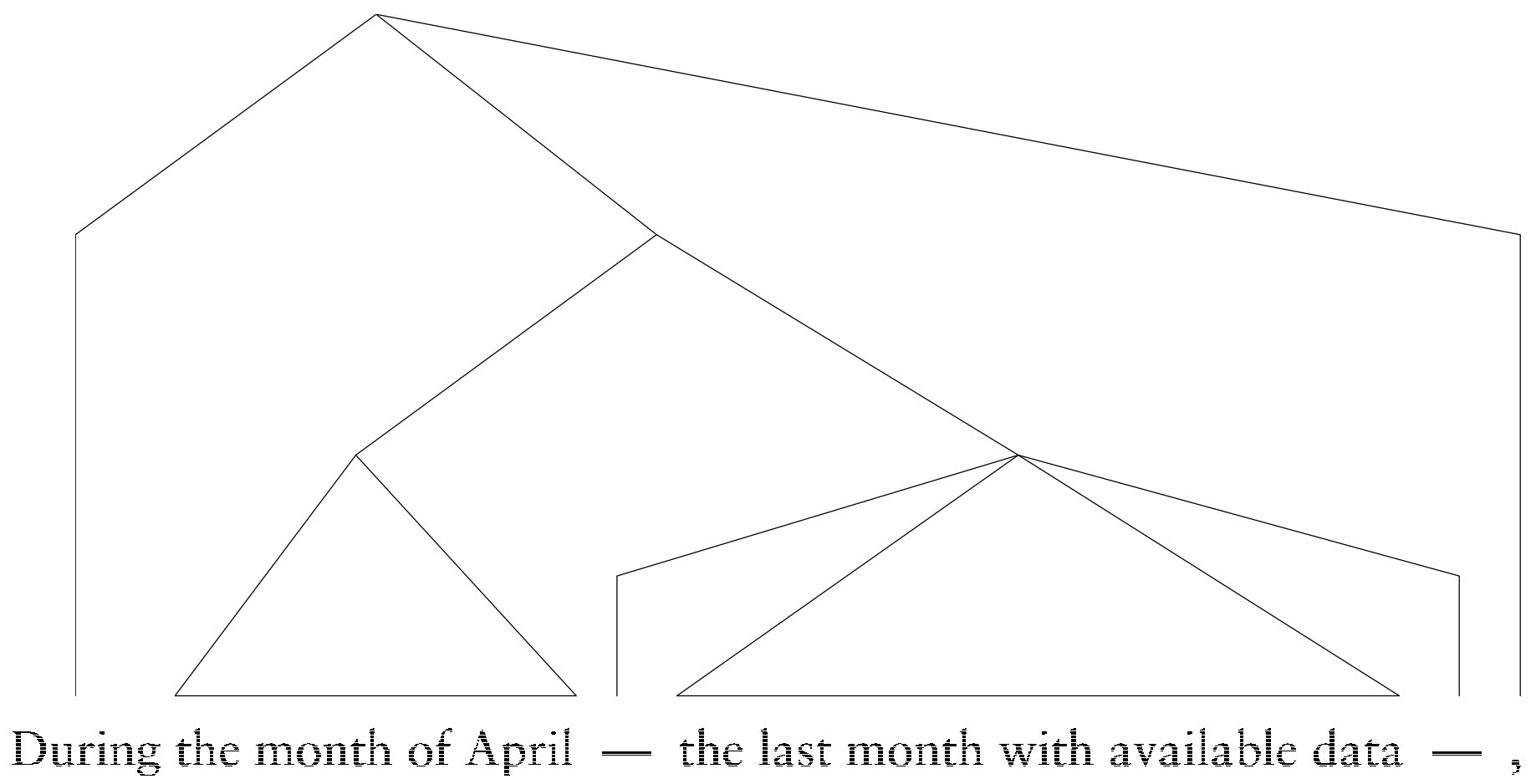

**Figure 3**

On the one hand, as in Briscoe's approach, the above sentence-initial PP is not assumed to have underlyingly balanced commas; this obviates the need to make sure a comma is not presented before the first word of the sentence, as would be required in Nunberg's account. On the other hand, as in Nunberg's approach, it is the point absorption rule which is responsible for making sure a comma is not presented after the dash; because this rule operates at the string level, this obviates the need for a complicated featural treatment in syntax, as would be required in Briscoe's account.[4]

When considered in further detail, this case reveals an interesting contrast between the parsing and generation perspectives. From the parsing perspective, the easiest way to make sure (7) parses would be to make commas optional after sentence-initial PPs. The drawback of doing so would be that the grammar over-generates, yielding (8) as mistakenly grammatical:

(8) * During the month of April—the last month with available data—, the preliminary design of Project Reporter continued.

While this prospect is rather harmless from the parsing perspective, from a generation standpoint it is dreadful (or at least embarrasing).

Having argued the merits of a point absorption rule relying on adjacency rather than hierarchy, it remains to address Briscoe's criticism of Nunberg's account as (1) overly procedural and (2) requiring underlying forms not attested in text. With regard to this second item, as noted in the previous section, Nunberg is simply vague as to whether 'extra' points ever appear in intermediate structures. In the next section, I will show how point and bracket absorption rules can be incorporated into the insertion process, eliminating the need for such marks to appear explicitly.

[4] This point can be strengthened by observing that this dependency appears to be unbounded; cf.: *In the photo of the sketch of the painting of Mary—my dear old friend from freshman year—there were three clues to the mystery.*

Turning now to the first item, although Nunberg's discussion of rule orderings does come across as highly procedural, he does suggest that in principle there should be no need to stipulate these orderings (p. 72):

> This leaves us with a scheme of rule-ordering that mirrors the general ordering properties—and no less interesting, the complexity of interaction—that we observe in the phonological systems of spoken language. The earliest rules are those that make reference to abstract syntactic features. Subsequent rules make reference to indicator types (i.e., commas, semicolons, periods), and then to indicators considered as graphical elements. Thus we infer that no explicit statements about ordering need be included in the grammatical description of presentation rules.

Though boldly stated, Nunberg's conclusion that rule orderings need not be stipulated is not backed up with further technical details; nevertheless, the general direction he points to seems intuitively clear. As such, fleshing out this skeleton of an idea will be one of the main goals of the next section.

# 4. Towards a Stratificational Account

## *4.1 Ingredients of the Account*

The main ingredients of the proposed account are listed below.

### 4.1.1 Text realizer, sentence realizer and formatter

The account relies upon a syntactic realization architecture in which the sentence realizer is integrated with a text realizer and a formatter. The formatter's purpose is simply to insulate the text and sentence realizers from the idiosyncracies of the desired output markup language. The text realizer is more significant, in that it is responsible for passing some formatting specifications directly to the formatter: these include titles, sections, etc., as well as discourse-level brackets, bullets and enumerations. The pertinent properties of the sentence realizer are described in the remainder of this list.

### 4.1.2 Presentation rules

As mentioned in the preceding section, the present account incorporates the point and bracket absorption rules into the insertion process, eliminating the need for 'extra' marks to appear explicitly. This happens in two steps:

1. For each possible insertion site, the set of points vying for the insertion site is identified.

2. If the insertion site is within the scope of a bracket, no point is presented; otherwise, the strongest point is presented, where commas are considered the weakest points, periods the strongest, and dashes, semi-colons and colons fall inbetween, in that order.

This rule will be illustrated in the section 4.4; for present purposes, it suffices to note that by identifying the possible insertion sites for points and inserting at most one point in each site, this rule sidesteps Briscoe's criticism regarding unattested forms.

The remaining rules are simply listed by component next.

#### 4.1.3 Syntactic, morphological and graphical components

In order to properly account for the interaction of the various presentation rules, these rules must be stratified into three components. These components, which we may generically term the syntactic component, the morphological component, and the graphical component, focus on the properties of hierarchy, adjacency and graphical form, respectively.

Given these emphases, the most sensible way to allocate the requisite presentation rules is as follows:

- **Syntactic component:** This component contains rules for comma promotion, bracket and italics alternation, boundary propagation, and bracket insertion. Note that boundary propagation is simply the transmission of features of a phrase to the boundary elements of the phrase; e.g., that a phrase marked as between parentheses must have a left parenthesis to the left of the leftmost element, and a right parenthesis to the right of the rightmost element. Also note that bracket insertion must be performed here, as the scope of brackets is vital to preserve.

- **Morphological component:** The rule for converting features to points (as described above) is contained in this component. The location of the point insertion rule in this component reflects Nunberg's generalization concerning point and bracket absorption, i.e. that these phenomena are sensitive to adjacency rather than hierarchy.

- **Graphical component:** Rules for quote transposition, graphic absorption, and font- and face-harmony are located in the last component. This reflects the intuition that these rules are aesthetically motivated, as Nunberg observes (cf. pp. 77-79). Note that in order to achieve the desired results, quote transposition must be blocked in exceptional circumstances, and font- and face-harmony must be applied to matching pairs of brackets.

#### 4.1.4 Visual structures

As noted in section 2.1.4, distinctions of font and face should be kept separate from brackets, points and formatting devices such as bullets and numbers, in order to eliminate problems in determining adjacency. Consequently, visual structures are employed which are orthogonal to the main representational structures employed in the above components. Separating out font and face in this way reflects how these formatting devices differ from those that are realized as independent written marks.

### *4.2 Derivative Phenomena*

Stratifying the presentation of punctuation in the manner just described suffices to account for the phenomena listed below in a derivative (rather than explicit) fashion.

#### 4.2.1 Ordering of absorptions and transpositions

In section 2.1, we reviewed Nunberg's examples showing that comma promotion should precede point absorption, which in turn should precede quote transposition, which should then precede graphic absorption. Closer examination of the cases discussed in section 2.1.4 reveals that quote transposition and graphic absorption could instead take place opportunistically, reducing this ordering to a partial one. As such, Nunberg's observed ordering turns out to be consistent with the ordering induced by the stratification of rules into components described in the preceding section: since the comma promotion rule applies in syntax, it must apply before the point insertion rule (where point absorption takes place); similarly, since the point insertion rule takes place in morphology, it must apply before either the quote transposition or the graphic absorption rule.

Now, while it might seem that an explicit ordering between rules has simply been replaced with an explicit ordering between components, this need not be the case. Formally, the presentation rules can be allowed to apply in any order, as long as they are restricted to apply to structures of the right type: in other words, the desired ordering can be achieved via the type discipline used to organize the components, rather than by stipulation. For example, one way to accomplish this goal would be to let the syntactic component contain rules which map between structures of type A and B, the morphological component rules which map between structures of type B and C, and the graphical component rules which map between structures of type C and D. Given this setup, the desired ordering among phenomena will result regardless of whether the rules are applied in a pipelined or opportunistic fashion.

#### 4.2.2 Abbreviation periods

As shown in section 2.1.4, abbreviation periods are not subject to point absorption. Assuming that these periods are introduced lexically, rather than through syntactic features, the point absorption rule sketched above will not apply, as desired.

### 4.2.3 Discourse-level brackets and vertical lists

In section 2.3, we saw that bracket absorption does not apply when brackets enclose one or more complete sentences, and that comma promotion applies to vertical lists which form a single sentence. By letting the text realizer bypass the sentence realizer in the case of discourse-level brackets, the bracket absorption rule can be naturally circumvented; conversely, by not bypassing the sentence realizer in the case of vertical lists forming a single sentence, comma promotion will apply as usual.

### 4.2.4 Transparency of font and face

Keeping font and face in a separate structure avoids the artificial problem of determining adjacency in the presence of formatting directives, as was illustrated with example (1), repeated below.

(1) `*After she wrote :BEG-ITAL What Next? :END-ITAL .`

## *4.3 The Account in MTT*

For the most part, the present account has been instantiated within the framework of the CoGenTex core realizer, which is largely based upon Meaning-Text Theory (henceforth MTT; Mel'čuk and Pertsov, 1987; Mel'čuk, 1988). Work on the implementation is also in progress, and nearing completion (cf. Lavoie, 1995). Details of the rules for the alternations and promotions discussed in section 2.2, however, have yet to be finalized.

In MTT, a succession of levels are defined, as well as a succession of components which map between these levels. The most abstract syntactic level is the Deep-Syntactic Representation (DSyntR), a dependency tree representing the syntactic relationships amongst the meaning-bearing lexemes of a sentence. The next level, the Surface-Syntactic Representation (SSyntR), is also a dependency tree, this time including all lexemes of the sentence and the surface-syntactic relations amongst them. Following the SSyntR is the Deep-Morphological Representation (DMorphR), where the nodes of the SSyntR are linearized; subsequent to the DMorphR is the Surface-Morphological Representation (SMorphR), where lexemes labeled with morphological features have been mapped to actual morphemes. The final two levels (on the written language side) are the Deep- and Surface-Graphical Representations (DGraphR, SGraphR, resp.).

In each of these levels, the representation is divided into a main and various auxiliary structures. Of particular interest here will be the written-language analogues of the Morphological-Prosodic Structures, which we will call Morphological-Visual Structures (MorphVisSs).

The highly stratified nature of MTT facilitates the instantiation of the present account within the framework of the CoGenTex core realizer, as follows:

1. Since the text realizer and formatter have been independently motivated and developed (for the most part), few changes are required here.

2. The presentation rules, many of which are illustrated in the next subsection, are integrated into the components listed below (as described in the preceding subsection).

3. The requisite syntactic component in MTT is the Surface-Syntactic Component, i.e. the component which maps SSyntRs to DMorphRs. The requisite morphological component is the Deep-Morphological Component, i.e. the component which maps DMorphRs to SMorphRs. Finally, the requisite graphical component is the Surface-Morphological Component, i.e. the component which maps SMorphRs to DGraphRs. (Note that the SGraphR is simply an application-independent markup language which serves as input to the formatter.)

4. The requisite visual structures (DSyntVisS, SSyntVisS, DMorphVisS, SMorphVisS, and DGraphVisS) have been introduced as analogues to their spoken-language counterparts.

### *4.4 A Worked Example*

To demonstrate the account in action, we will now work through example (9) below in some detail:

(9) (Three programmers—including “Mr. Q.A.,” from CoGenTex—will work on *Project X.Y.Z.*)

This text is considered to consist of an independent sentence surrounded by parentheses. The (abbreviated) text structure for this example is shown in Figure 4 below:

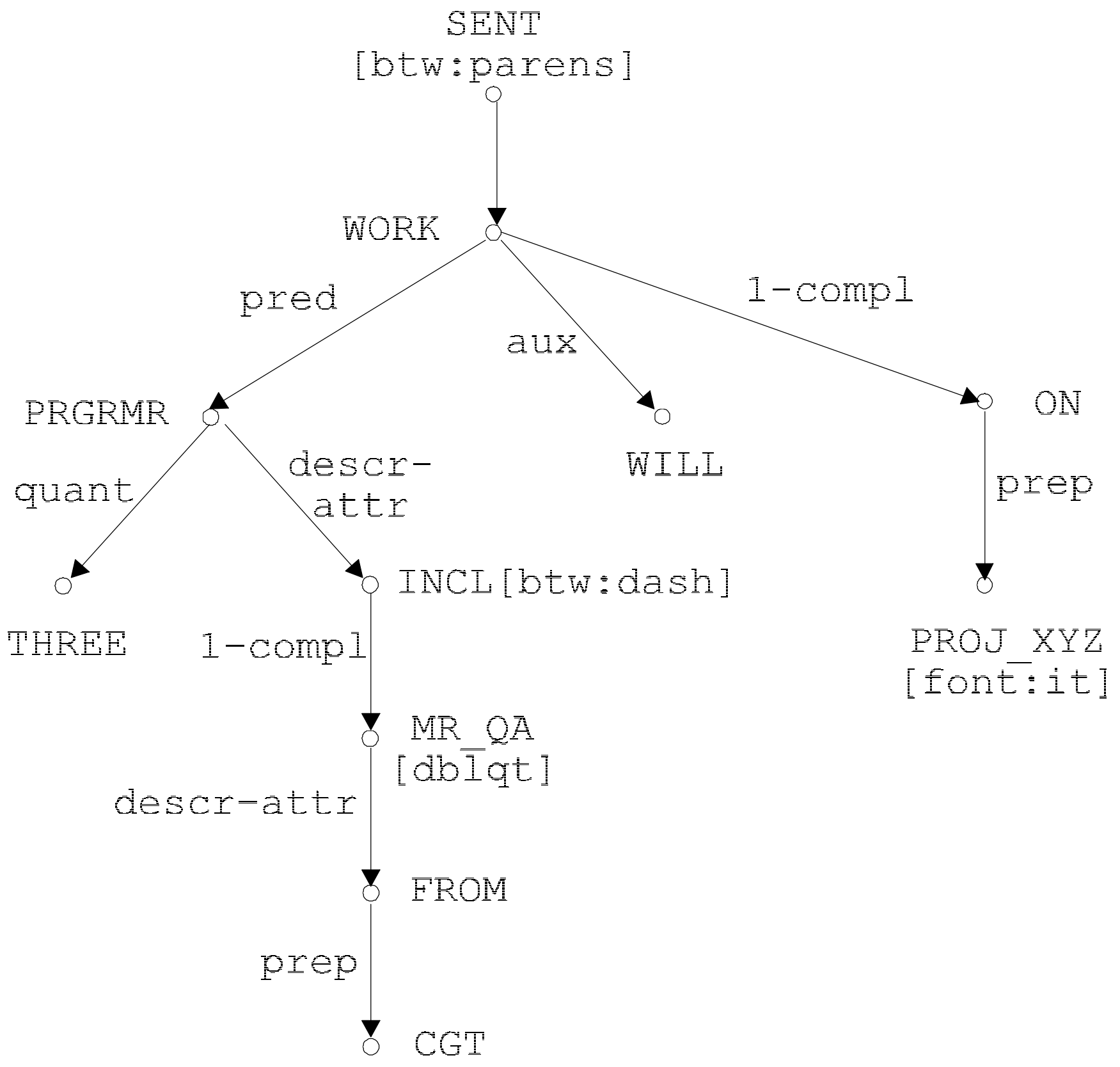

**Figure 4**

Here we have a sentence node, which is marked as being between parentheses, which points to the SSyntR for this sentence.[5] The main structure of the SSyntR is a dependency tree, whose nodes are labeled with lexemes and whose edges are labeled with surface syntactic relations. The various auxiliary structures are implemented as separate feature lists; only the SSyntVisS is shown here. Note that these attributes may be either phrasal or lexical. For example, in Figure 4, `[dblqt]` and `[font:it]` are lexical attributes, and thus apply only to the lexemes `MR_QA` and `PROJ_XYZ`, respectively; in contrast, `[btw:dash]` is a phrasal attribute, and thus applies to the entire phrase headed by `INCL`. (Quotes, italics, and other similar formatting devices can also give rise to phrasal attributes, in which case they are notated `[btw:dblqt]`, `[btw:font:dash]` and so on.) Also note that whereas some punctuation and formatting features appear explicitly, other are introduced by default; for instance, here the commas delimiting *from CoGenTex* are triggered by the descriptive-attributive relation, according to the following rule:

(10)
```
X --descr-attr--> Y[~btw:dash]
      <-->
X + Y[btw:comma]
```

[5] N.B.: In the actual implementation, it would point to the DSyntR.

This rule simply states that if Y is a dependent of X via the descriptive-attributive relation, and Y is not marked as appearing between dashes, then Y should be linearized after X and between commas.

The DMorphR for this example is shown below:

(11)
```
THREE PRGRMR INCL[left:dash] LT-DBLQT MR_QA RT-DBLQT
FROM[left:comma] CGT[right:comma,right:dash] WILL WORK
ON PROJ_XYZ[font:ital,right:pd]
```

Here the nodes of the SSyntR have been linearized, the punctuation and formatting features propagated, and the quotes inserted. This establishes the insertion sites for the various points, which appear in the SMorphR:

(12)
```
Three programmers DASH including LT-DBLQT Mr. Q.A. RT-
DBLQT COMMA from CoGenTex DASH will work on
Project[begin:font:ital] X.Y.Z.[end:font:ital] PD
```

Note that the comma following `CGT` has been absorbed, since the dash takes precedence for this insertion site; also, the comma preceding `FROM` is not absorbed via point absorption or bracket absorption, since quote transposition has yet to apply. The final level of interest, the DGraphR, appears below:

(13) Three programmers—including “Mr. Q.A.,” from CoGenTex—will work on *Project X.Y.Z.*

At this stage quote transposition and graphic absorption have applied. To complete the realization, all that remains is to insert the discourse-level parentheses around this text.

## 5. Summary and Conclusions

In this paper, I have examined Nunberg’s (1990) approach to presenting punctuation, and suggested that if not applied overzealously, his approach promises to considerably simplify the syntactic realization of punctuation and other formatting devices. I have also developed a stratificational account which elaborates and improves upon aspects of Nunberg’s; in particular, I have proposed an improved treatment of point and bracket absorption, as well as fleshed out his ideas as to how the need to stipulate rule orderings might be avoided. Finally, I have also sketched how the present account can be instantiated within the Meaning-Text framework of the CoGenTex core realizer.

## Acknowledgements

This research was partially supported by USAF Rome Laboratory SBIR contract F30602-94-C-0124.

For valuable discussion, criticism and insights, I would like to thank Ted Briscoe, Christy Doran, Lidija Iordanskaja, Tanya Korelsky, Benoit Lavoie, Igor Mel'čuk, Daryl McCullough, Owen Rambow, Ehud Reiter, and the three anonymous reviewers. For encouragement, a special debt is owed to Ehud Reiter, Tanya Korelsky and Igor Mel'čuk. Technical advice and many of the details of the MTT-based instantiation of the account were provided by Benoit Lavoie. All remaining mistakes are courtesy of the author.